\documentclass[11pt]{article}
	
	\newcommand{\blind}{0}
	
    \makeatletter
    \renewcommand\section{\@startsection {section}{1}{\z@}%
                                       {-3.5ex \@plus -1ex \@minus -.2ex}%
                                       {2.3ex \@plus.2ex}%
                                       {\normalfont\fontfamily{phv}\fontsize{14}{19}\bfseries}}
    \renewcommand\subsection{\@startsection{subsection}{2}{\z@}%
                                         {-3.25ex\@plus -1ex \@minus -.2ex}%
                                         {1.5ex \@plus .2ex}%
                                         {\normalfont\fontfamily{phv}\fontsize{12}{17}\bfseries}}
    \renewcommand\subsubsection{\@startsection{subsubsection}{3}{\z@}%
                                        {-3.25ex\@plus -1ex \@minus -.2ex}%
                                         {1.5ex \@plus .2ex}%
                                         {\normalfont\normalsize\fontfamily{phv}\fontsize{12}{17}\selectfont}}
    \makeatother
	
	\usepackage{amsmath}
	\usepackage{graphicx}
	\usepackage{enumerate}
	\usepackage{natbib} 
	\usepackage{url} 
	
\begin{document}
		
		\def\spacingset#1{\renewcommand{\baselinestretch}%
			{#1}\small\normalsize} \spacingset{1}
		
		\if0\blind
		{
		\title{AI perspectives in Smart Cities and Communities to enable road vehicle automation and smart traffic control}
			\author{Cristofer Englund$^{1,2,\ast}$, Eren Erdal Aksoy$^1$, Fernando Alonso-Fernandez$^1$, \\ Martin Daniel Cooney$^1$, Sepideh Pashami$^1$,  and Bj\"{o}rn \r{A}strand$^1$  \\
			\\
			\\
			\small{$^1$Center for Applied Intelligent Systems Research, Halmstad University, Sweden} \\
            \small{$^2$RISE Research Institutes of Sweden,  G\"{o}teborg, Sweden}\\
            \small{$^\ast$Corresponding author: cristofer.englund@ri.se} }
			\date{}
			\maketitle
		} \fi
 
		\if1\blind
		{

            \title{\bf \emph{IISE Transactions} \LaTeX \ Template}
			\author{Author information is purposely removed for double-blind review}
			
\bigskip
			\bigskip
			\bigskip
			\begin{center}
				{\LARGE\bf \emph{IISE Transactions} \LaTeX \ Template}
			\end{center}
			\medskip
		} \fi
		\bigskip
		
	\begin{abstract}
Smart Cities and Communities (SCC) constitute a new paradigm in urban development. SCC ideates on a data-centered society aiming at improving efficiency by automating and optimizing activities and utilities. Information and communication technology along with internet of things enables data collection and with the help of artificial intelligence (AI) situation awareness can be obtained to feed the SCC actors with enriched knowledge. This paper describes AI perspectives in SCC and gives an overview of AI-based technologies used in traffic to enable road vehicle automation and smart traffic control. Perception, Smart Traffic Control and Driver Modelling are described along with open research challenges and standardization to help introduce advanced driver assistance systems and automated vehicle functionality in traffic. To fully realize the potential of SCC, to create a holistic view on a city level, the availability of data from different stakeholders is need. Further, though AI technologies provide accurate predictions and classifications there is an ambiguity regarding the correctness of their outputs. This can make it difficult for the human operator to trust the system. Today there are no methods that can be used to match function requirements with the level of detail in data annotation in order to train an accurate model. Another challenge related to trust is explainability, while the models have difficulties explaining how they come to a certain conclusions it is difficult for humans to trust it.  
 	\end{abstract}
			
	\noindent%
	
	\spacingset{1.5} 

\section{Introduction}
 Smart Cities and Communities (SCC) is an emerging research field that spans in many dimensions and is promoted by major advances in technology, changes in business operation, and the overall environmental challenge. This paper reviews and positions the authors research domain and interest within AI in Smart Cities while also pointing out the research challenges for future research. One of the enabling technologies for SCC is Information and Communication Technologies (ICT) that connects infrastructure, resources, and services with data surveillance and asset management systems. Another technology is Internet of Things (IoT), which enables even the smallest device to connect to internet and share its operational status~\cite{smartcities3030052, scuotto2016internet}. Devices from e.g. the transportation system, power plants, or residential houses can be connected with the use of IoT technology~\cite{arasteh2016iot}. Modern business environments are highly competitive, and organizations are constantly finding ways to reduce cost. In addition, businesses are exploring new ways of developing their operations, and conversion towards not only providing a product but also a service connected to the product, which is becoming popular in many domains~\citep{tukker2015product}. Economics can be seen as the main driver for industry, and the environmental challenge as the main driver for political and private actors~\citep{garling2007travel}. We have a great responsibility to protect our natural resources for our descendants. 

To set free the potential of SCC, in any application area, the collected data needs to be processed and analyzed. With the help of AI, relationships, root causes, and patterns can be found in the data. AI can then use the new information to tailor guidance and provide suggestions to users on how to improve behavior~\citep{byttner2011consensus,englund2008ink}. 

There, still, exist various challenges related to SCC, some of which are listed in the next section. 



\subsection*{Challenges addressed by SCC}

Business operation has changed dramatically during the last 60 years. GDP changes from 1947-2009 clearly show a decrease in industry and growth in professional and business services. In the US, the GDP decrease in industry is around 50\%. On the other hand, the growth in GDP for professional and business services is 400\%~\citep{gdp1947}. 
This trend is holding on and in the last 17 years the GDP service share has increased from 72.8\% to 77.4\% of GDP whereas industry share has decreased from 22.5\% to 18.2\%~\citep{Statista}.

Digitalization, with help from ICT, is a major contribution to this trend and with the latest technology trends in computing power, AI is becoming a key technology to make use of the data to further develop the services. 

Another challenge is energy consumption and in particular energy that comes from non-renewable sources such as oil. In Europe, the primary energy consumption was 1 561 million tonnes of oil equivalent (Mtoe) in 2017, 5.3\% above the EU target for 2020. In 2018 energy came from petroleum products including crude oil (36\%), natural gas (21\%)  renewable energy (15\%), solid fossil fuel (15\%), nuclear energy (13\%) ~\citep{Eurostat2020}.  The energy consumption by sector in the EU breaks down in the following way: the industry sector (31\% of final energy consumption), the transport sector (28\%), households (25\%), services (13\%) and agriculture \& forestry (2\%)~\citep{Eurostat2019}. 

In this paper, we primarily address transportation within SCC and how AI can be used to improve efficiency and thus, reduce the energy consumption. AI can be used to learn traffic behavior and to control traffic both on the micro level in e.g. intersections~\citep{Chen2016a} and on the macro level~\citep{englund2014cooperative}. 

In the 28 EU Member States (EU-28) energy consumption in transport has increased with 32\% between 1990-2017. In the EU-13 states, the increase is 102\% during the same period. Road transport accounts for 73\% of the total energy consumption in the transport sector, and road transport alone has increased 34\% between 1990-2017~\citep{EEA2019}. 

In 2020, the expectation due to the Covid-19 pandemic, is that the energy demand is 10\% below the 2019 levels. This would be twice the decline experienced throughout the financial crisis 2008-2009. CO2 emissions in the EU declined by 8\% during the first quarter of 2020 compared with the same period in 2019~\citep{EU2020}.

 Traffic safety is also a global challenge and traffic accidents have become one of the most common causes of death among young people~\citep{WHO2015}. Although fatalities have decreased for motorists in most countries, this is not the case for vulnerable road users (VRUs),~\citep{niska2013statistik} including pedestrians, bicyclists and moped riders. In Europe 22 700 people lost their lives in 2019~\citep{EURoadSafety2019} and more than 1.4 million people were injured in 2018~\citep{EURoadSafety2018}. Worldwide, 1.35 million people lost their lives and up to 50 million were injured in traffic accidents in 2018~\citep{WHOSafety2018}. 
 
 Given these facts about business development, energy and traffic safety, it is clear that there is huge potential in society for improvements, both in terms of energy efficiency and traffic safety. 
 
 With a starting point in SSC, the enabling technologies and the challenges mentioned above, this paper will focus on how AI can be used to enable energy saving and improved traffic safety within SCC. Consequently, the EU has set goals on energy consumption in the transportation sector. By 2030 the climate and energy goals for a competitive, secure and low-carbon EU economy, state that the greenhouse gas (GHG) emissions should be reduced by 55\% below the 1990 level and the amount of renewable energy should be at least 27\%~\citep{EuClimate,Commission2014}. 

 With the trend towards increased vehicle automation, there is a large potential for reducing the effects of an accident, or, if possible, avoiding the accident completely. This can be done by IoT; i.e. building sensor-based safety systems that can detect VRUs and give warnings or actively react on the information. Enabling the development of such systems requires knowledge of how these road users behave, and how that behavior can be described so that the automated vehicle functions can make correct interpretations and decisions. Also, vehicular communication can help traffic coordination and reduce travel time for e.g. emergency vehicles~\citep{barrachina2014reducing, englund2016grand,englund2014cooperative}. Consequently, the European commission has set goals on traffic safety, i.e. close to zero deaths in 2050 and an interim goal, halving the number of seriously injured by 2030 from the level of 2020~\citep{Commission2019}.

Energy efficiency along with traffic safety are the two main goals of SSC in the domain of the transportation system. 

 With the help of AI and data analytics it may be possible to improve utilization of the manageable assets within the transportation system. In particular, this paper describes the on-board AI-based systems along with the infrastructure AI-based systems that build up SSC addressing traffic safety and efficiency. An overview is given for the research areas of perception, traffic control, and interaction.

The rest of the paper is organized as follows. Section~\ref{sec:initiatives} describes research initiatives, projects and financial programs. Section~\ref{sec:approaches} describes the different approaches of using AI in SCC such as perception, traffic system control, and driver monitoring. Section~\ref{sec:open_research} highlights open research questions and standardization to facilitate implementation and adoption. Finally, Section~\ref{sec:conclusion} provides a summary and conclusion of the findings.

\section{Research initiatives within Smart Cities and Communities}\label{sec:initiatives}

In the EU, the European Innovation Partnership on Smart Cities and Communities (EIP-SCC) is an initiative supported by the European Commission that brings together cities, industry, Small and medium-sized enterprises (SMEs), banks, research and other smart city actors to share information and find partners for projects~\citep{EU-smartCities}.

The EU project CITYKeys is funded by the European Union HORIZON 2020 programme. In collaboration with cities, the project developed and validated key performance indicators along with data collection procedures for common and transparent monitoring to be able to compare smart city solutions across European cities~\citep{CityKeys}. The project has divided the Smart City into sub themes that concern e.g. diversity and social cohesion and aim at promoting diversity, community engagement and social cohesion within a community. Education focuses on improving accessibility and quality of education for everyone. Safety concerns lowering the rate of crime and accidents. Health focuses on improving quality and accessibility of public health systems, for everyone, as well as encouraging a healthy lifestyle. Quality of housing and the built environment promote development of mixed-income areas, ensure high quality and quantity of public spaces and recreational areas, and improve affordability and accessibility to good housing for everyone. Finally, quality of life also promote access to (other) services that focus on providing better access to amenities and affordable services in physical and virtual space for everyone. CITYKeys also aims to harmonize data collection from the involved cities that can enable comparisons of the performance of the measures introduced to reach the EU energy and climate targets.

CIVITAS~\citep{EU-scivitas} is a pan European network of cities that are dedicated to cleaner and better transports. The network is financed by the European Commission and was launched in 2002. Since then, 85 cities have joined the network and more than 900 measures and urban transport solutions have been tested and implemented. The main goal of CIVITAS is to make it easier for cities to obtain cleaner and better connected transport solutions in Europe and beyond. The four main characteristics of CIVITAS are a living lab approach to bring out research projects, maintaining a network of cities for cities, facilitating a public private partnership, and promoting political commitment. The network gives unique opportunities for practitioners to experience innovative transport solutions and learn from experts in the field. Sustainable mobility is the overall area and the project is divided into 10 sub areas: Car-Independent Lifestyles; Clean Fuels \& Vehicles; Collective Passenger Transport; Demand Management Strategies; Integrated Planning; Mobility Management; Public Involvement; Safety \& Security; Transport Telematics; Urban Freight Logistics.

In Sweden where this research is brought out, there are a number of initiatives from the Swedish government to support the development of SCC. The Strategic Innovation Program (SIP) Drive Sweden is a Swedish cross disciplinary collaboration platform driving the development towards sustainable mobility solutions for people and goods~\citep{drive_sweden}. Drive Sweden is an important stakeholder in future mobility that fertilizes national as well as international collaboration to encourage the development of future sustainable mobility. Drive Sweden is financed by the Swedish Innovation Agency, Vinnova, the Swedish Energy Agency and the Swedish research council for sustainable development, Formas. Drive Sweden also provides a weekly newsletter that summarize news within the area of smart mobility~\citep{DriveSwedenNewsletter}.
Examples of projects that has been financed by Drive Sweden, within the field of future mobility, are for example: \emph{Study of communication needs in interaction between trucks and surrounding traffic in platooning}, \emph{Intelligent and self-learning traffic control with 3D \& AI} and \emph{Security for autonomous vehicles from a societal and safety perspective
}.

InfraSweden2030~\citep{infraSweden} is another Strategic Innovation Program in Sweden. Whereas Drive Sweden focuses on future mobility, InfraSweden2030 focuses on the transportation infrastructure of the future. InfraSweden2030 is also financed by Vinnova, the Swedish Energy Agency and Formas. The aim of the program is to contribute to reduced climate and environmental impact from the construction, operation and maintenance of the transport infrastructure. The programme organizes seminars and workshops to facilitate collaboration and innovation within the Swedish transport infrastructure sector in order to address the society's economic and social challenges. In addition, InfraSweden2030 funds research projects that address these goals and challenges. The three objectives of the project are Develop innovation for transport infrastructure, create an open, dynamic and attractive environment, and Reduce impacts on the environment and climate. An example of a project financed by the InfraSweden2030 programme is the \emph{iBridge} project. The overall aim of the project is to automate and make available knowledge about bridges that can be used to lower maintenance cost.


Viable Cities~\citep{ViableCities} is also a Swedish Strategic Innovation Program. Viable Cities focuses on smart sustainable cities. The vision of the programme is to accelerate the transition towards inclusive, climate neutral cities until 2030 with a good life for all with the help of digitalization and citizen engagement. Viable Cities is like its siblings Drive Sweden and InfraSweden2030 financed by Vinnova, Swedish Energy Agency and Formas. 
Within the Viable Cities programme, a strategic initiative, Viable Cities Transition Lab is formed to foster capabilities to handle the societal challenges within climate and environment transition. The Transition Lab aims at luring the full potential out of humans in the era of digitalization and automation--thus, obtaining new methods to transform society for an equal and circular economy, and responsible and ground-breaking technology to create behavioral change to a more sustainable and entrepreneurial society. \emph{Xplorion - Residential mobility service in car-free accommodation} is a project financed by Viable Cities~\citep{Xplorion}. It offers a mix of mobility services such as public transport, carpool and bicycle pool, to households in a new residential area called Södra Brunnshög in Lund Sweden. The aim is to provide mobility as part of the residential rent and thus, allow more efficient use of transport leading to a reduction in emissions from residents' travel. It is also expected that by connecting housing and mobility, there will be synergies that will make the resources in the system more efficient than today.

Smart City Sweden~\citep{SmartCitySweden} is a governmental export platform for sustainable city solutions. The platform reaches out to international delegates who are interested in investing in smart \& sustainable city solutions from Sweden. The platform has five focus areas: Mobility; Climate, Energy \& Environment; Digitalization; Social Sustainability; and Urban Planning. Through their web page Smart City Sweden promotes solutions ranging from smart production of biogas from household waste, and water treatment facilities to congestion pricing solutions in Stockholm, future multi-modal transportation services and service platforms aiming at supporting an automated transportation system.

\section{Approaches}\label{sec:approaches}


\begin{figure}[!t]
  \includegraphics[trim=0 0 10 0,clip,width=\linewidth]
   {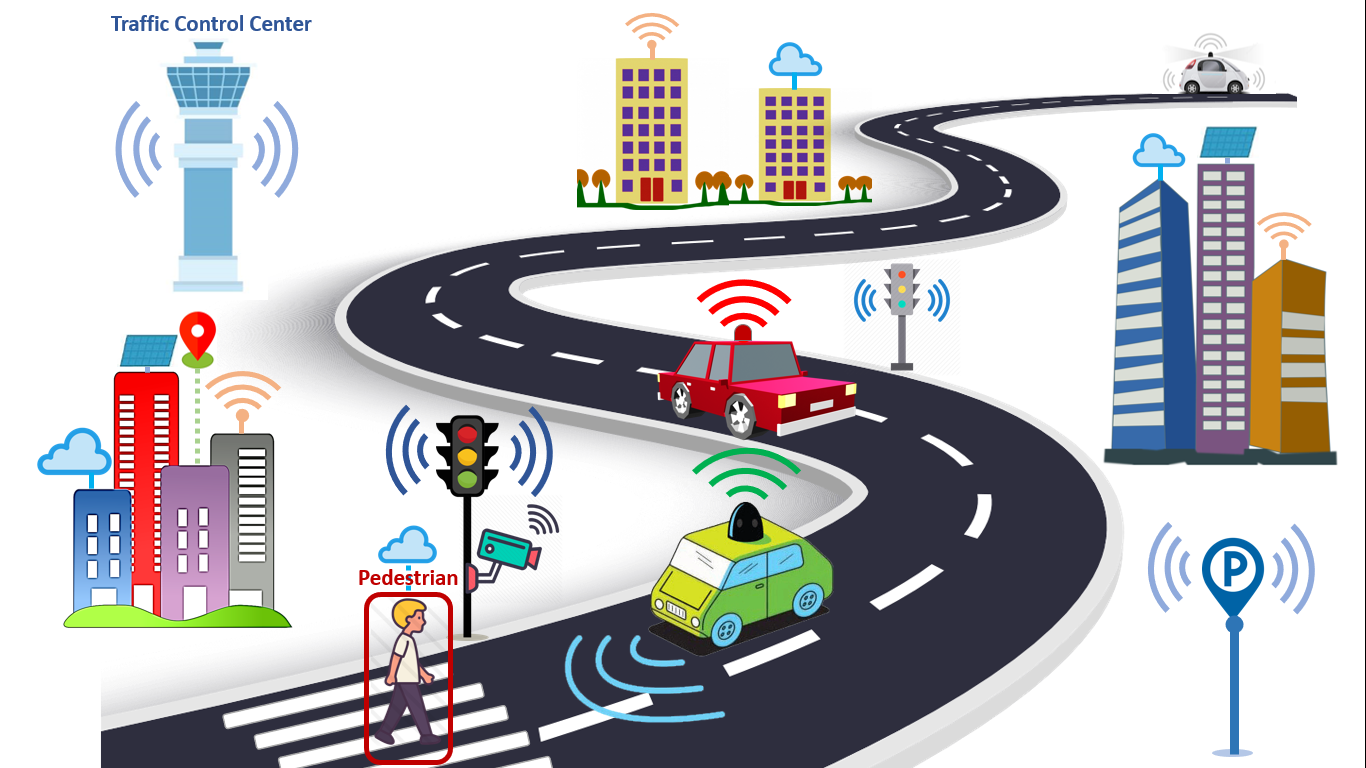}
   \caption{A sample smart city scenario. Information is continuously shared among different buildings, infrastructures and vehicles. Thus, autonomously operating vehicles safely react to detected VRUs.}
   \label{fig:smartcity}
\end{figure}

 \begin{table}[!b]
     \centering
     \begin{tabular}{lccc}
          &Strategical&Tactical&Operational  \\
          \hline
         In-vehicle & x&x&x\\
         Infrastructure &x &x&\\
         \hline
     \end{tabular}
     \caption{Overview of in-vehicle and infrastructure-based systems' contribution to road vehicle automation. }
     \label{tab:SOT}
 \end{table}

This section describes different approaches of using AI in SCC. The paper is founded from the perspectives of the authors own research, the paper initially gathers previous research in the SCC domain and secondly, in regards to operational, tactical and strategical vehicle and traffic functions, the paper describes future challenges from a speculative design approach to enable road vehicle automation and smart traffic control.

 Figure~\ref{fig:smartcity} illustrates a sample smart city scenario where information is continuously shared among different units such as smart buildings, vehicles and infrastructures to enable road vehicle automation and smart traffic control applications. In Figure~\ref{fig:smartcity} a sample of a smart city is illustrated. Information is continuously shared among different buildings, infrastructures and vehicles. Thus, autonomously operating vehicles safely react to detected VRUs.

Table~\ref{tab:SOT} shows how in-vehicle and infrastructure-based systems contribute to the different levels of control in automated driving. The driving tasks can broadly be categorized into three different levels {\em i.e.} strategic, tactical and operational ~\citep{Aramrattana2015}. The {\em strategic} tasks comprise high-level (and longer-term) planning decisions such as route choice, traffic flow control, and fuel cost estimates etc., whereas {\em operational} tasks include low-level (short-term) and continuous routine tasks such as lateral control based on immediate environmental input, and in-vehicle input such as driver monitoring. The {\em tactical} tasks fall in the middle of the two and are mid-level, medium-term tasks including, but not limited to, turning, overtaking, gap adjustment, merging etc., based on local awareness around the vehicle. 
In the following subsections we describe {\em Perception systems} enabling situation awareness for autonomous vehicles, and approaches for {\em Traffic system control} and finally we give examples of {\em Driver monitoring} systems.

\subsection{Perception}\label{sec:perception}

Mobility within SCC concerns several of the challenges described above, e.g. traffic safety and environmental impact. These challenges in turn drive the technology development towards improved sensor systems that can improve  vehicles' perception to help the driver in hazardous situations with e.g. advanced driver assistance systems (ADAS). ADAS are functions that automate vehicle functions to improve safety or comfort. Examples of such functions are lane keeping aid, automated emergency braking, adaptive cruise control. Nevertheless, with the help of sensors and AI the vehicles' perception system is becoming more and more intelligent and there are now several examples of highly automated vehicular systems. Realizing road vehicle automation builds on the assumption that the vehicle can maneuver automatically by itself. This requires local awareness around the vehicle to be able to handle obstacles, hazardous situations, and unanticipated events. One way to achieve local awareness is through on-board sensors. Camera, radar, and LiDAR (light detection and ranging) sensor signals are typically fused to obtain scene understanding~\citep{sivaraman2013looking}. Such sensors operate in the range from a few centimeters to 200 m~\citep{kocic2018sensors}. 

Another way to obtain awareness in traffic is to use sensors in the infrastructure~\citep{Englund2020a,Englund2020b} and using wireless communication to exchange information between vehicles and infrastructure~\citep{lidstrom2009act}. This section describes perception systems enabling situation awareness in traffic.


Scene understanding is an essential prerequisite for autonomous vehicles to increase the local awareness. Semantic segmentation and object detection are two fundamental early lower-level perception components which help in gaining a rich understanding of the scene.
Safety-critical systems, such as highly automated vehicles, however, require not only highly accurate but also reliable predictions with a consistent measure of uncertainty. This is because the quantitative uncertainty measures can particularly be propagated to the subsequent units, such as decision-making modules that lead to safe maneuver planning or emergency braking, which is of utmost importance in safety-critical systems.
Therefore, semantic segmentation and object detection integrated with reliable confidence estimates can significantly reinforce the concept of safe mobility within SCC.

Given an image or point cloud data stream, there exist two mainstream deep learning-based AI approaches used for real-time object detection tasks \citep{Liu2020}: two-stage and one-stage detection frameworks. Two-stage methods \citep{Girshick2014,Girshick2015,He2017,Ren2017} initially have a preprocessing step to generate category-independent object region proposals. The output of this step is then passed to the category-specific classifier, which returns the category label for each detected proposal. On the other hand, one-stage detectors \citep{Szegedy2013,Redmon2016,ssd2016,Law2018} are region proposal-free frameworks where the proposal generation step is not separated, and thus the entire framework works in a unified end-to-end fashion. Such unified approaches directly predict class probabilities together with the bounding boxes using single feed-forward networks. In contrast to unified (one-stage) models, region-based (two-stage) methods achieve relatively better detection accuracies with the cost of being computationally more expensive. Unlike two-stage networks, detection accuracy of a one-stage model is less sensitive to false detections coming from the backbone network. 

Regarding the task of semantic scene segmentation, advanced deep neural networks are heavily used to generate accurate and reliable segmentation with real-time performance. Most of these approaches, however, rely on camera images~\citep{kendall2015bayesian,Zhao_2017_CVPR,Chen_2018_ECCV,deeplab_2018,FastSCNN2019}, whereas relatively fewer contributions have discussed the semantic segmentation of 3D LiDAR data~\citep{SqueezesegV01,rangenetpp,salsanext2020}. The main reason is that unlike camera images, LiDAR point clouds are relatively sparse, unstructured, and have non-uniform sampling although LiDAR scanners have wider field of view and return more accurate distance measurements.

As comprehensively described in~\citep{survey3DPC2019}, there exist two mainstream deep learning  approaches addressing the semantic segmentation of 3D LiDAR data only: pointwise and projection-based neural networks. The former approaches operate directly on the raw 3D points without requiring any pre-processing step~\citep{pointnet,pointnetpp,SPGraph}, whereas the latter project the point cloud into various formats such as 2D image view~\citep{rangenetpp,salsanet2019,salsanext2020,SqueezesegV01,SqueezesegV02} or high-dimensional volumetric representation \citep{Zhang2018,VoxelNet18}. There is, however, a clear split between these two approaches in terms of accuracy, runtime, and memory consumption. 
Projection-based approaches can achieve state-of-the-art accuracy while running significantly faster. Although point-wise networks have slightly fewer parameters, they cannot efficiently scale up to large point sets due to the limited processing capacity, thus, they have a longer runtime.

When it comes to uncertainty estimation, Bayesian Neural Networks (BNNs) are intensively used since such networks can  learn approximate distribution on the weights to further generate prediction confidences.
There exist two types of uncertainties: \textit{Aleatoric} which can quantify the intrinsic uncertainty coming from the observed data  and \textit{epistemic} where the model uncertainty is estimated by inferring with the posterior weight distribution, usually through Monte Carlo sampling. 
Unlike \textit{aleatoric} uncertainty which captures the irreducible noise in the data, \textit{epistemic} uncertainty can be reduced by gathering more training data.
For instance, segmenting out an object that has relatively fewer training samples in the dataset may lead to high \textit{epistemic} uncertainty whereas high \textit{aleatoric} uncertainty may instead occur on segment boundaries or distant and occluded objects due to noisy sensor readings which are inherent in the sensors. Bayesian modelling helps estimating both uncertainty types.

Gal \textit{et al.}~\citep{gal2016dropout} proved that dropout can be used as a Bayesian approximation to estimate the uncertainty in classification, regression, and reinforcement learning tasks while this idea was also extended to semantic segmentation of RGB images by Kendall \textit{et al.}~\citep{kendall2015bayesian}. 
Recently, both uncertainty types were applied to 3D point cloud object detection \citep{feng2018towards}, optical flow estimation \citep{ilg2018uncertainty}, and semantic segmentation of 3D LiDAR point cloud data \citep{salsanext2020}.


Typical algorithms for sensor fusion include Kalman filters~\citep{welch1995introduction}. Kalman filters are recursive filters that estimate the state of the system from several noisy measurements. This technology is use, for example, in~\citep{lidstrom2012modular} where a vehicle for the Grand Cooperative Driving Challenge was developed. Also in~\citep{kianfar2012design} Kalman filters were used for sensor fusion and scene understanding. 

For autonomous vehicles to behave efficiently and safely in traffic they require not only scene understanding derived from perceptual data, but also algorithms that can model and predict other road users’ behavior. Modelling of the behavior and prediction of motions has long been of interest and is applicable especially in domains where humans and intelligent systems co-exist \citep{RUDENKO-ARXIV-2019}. In a survey \citep{LEFEVRE-ROBOMECH-2014}, which addresses motion-prediction applications in intelligent vehicles, the authors proposed three main categories for how agent motion is modelled: physics-based, maneuver-based and interaction aware approaches. Work that focus on the acceleration and deceleration behavior of different vehicle types employ physics-based methods, e.g. \citep{BOKARE-WCTR-2016} and \citep{MAURYA-IJTT-2012}. \citep{LEFEVRE-REPORT-2013} suggest an interaction aware method for instance for risk assessment in traffic. Maneuver-based approaches assume that the maneuver intention can be recognized early on and future trajectory should match that maneuver. The main idea in this approach is that real-world trajectories from a road agent can be clustered into categories representing different behaviors. Based on a set of behaviors, maneuver-based motion prediction approaches employ estimation techniques, for instance Gaussian Processes~\citep{TAY-THESIS-2009, JOSEPH-AURO-2011} to then estimate most probable future maneuvers. Deep-learning techniques have also been applied to cluster vehicle encounters \citep{LI-ARXIV-2018}.

Roundabouts play a very important role in modern traffic infrastructure. Studies have shown that roundabouts reduce injury crashes (in comparison to signal-controlled intersections), can reduce delays and improve traffic flows, and even have lower long-term cost compared to signal-controlled intersections \citep{WSDT-WEBPAGE-2019}. A study that employs support vector machines to classify vehicles inside a roundabout to either stay or leave the roundabout is presented in \citep{ZHAO-IV-2017}. Similarly, a study to estimate the effects of the roundabout layout on driver behavior, employing simulation data is presented in \citep{ZHAO-ELECT-2017}. A method for estimating reachable paths using conditional transition maps is presented in \citep{KUCNER-IROS-2013}. A study that employs a stereo camera setup for time-to-contact estimation is presented in \citep{MUFFERT-IVS-2012}. This study is focused towards risk assessment instead of efficiency and smoothness of drive, when entering a roundabout. 

Recent work by Muhammad and Åstrand~\citep{muhammad2018intention} apply particle filters to predict road user behavior. In \citep{Muhammad2019} they address the problem of modelling and predicting agent behavior and state in a roundabout traffic scenario. They present three ways of modelling traffic in a roundabout based on ($i$) the roundabout geometry (which can be generated using drawings or satellite images, etc.); ($ii$) mean path taken by vehicles inside the roundabout; and ($iii$) a set of reference trajectories traversed by vehicles inside the roundabout. The roundabout models were compared in terms of exit-direction classification and state (i.e., position inside the roundabout) prediction of query vehicles inside the roundabout. The results show that the model based on a set of reference trajectories is better suited, in terms of both the early and robust exit-direction classification, and a more accurate state prediction. An additional experiment was done by categorizing vehicles into classes based on vehicle size (instead of a single class). Results indicate that such a categorization can affect, and in some cases enhance the state prediction accuracy. The particle filter approach in \citep{Magavi2020} was compared to a Recurrent Neural Network (RNN), namely, Long Short-Term Memory (LSTM) \citep{Gers1999} to determine the specific behavior model. Additionally, the network performance was compared with other RNN architectures such as the Bi-LSTM and Bi-LSTM + LSTM stacked architecture to evaluate which model has the best performance. Results showed a LSTM network can predict the exit of the vehicle in a roundabout much sooner than the particle filter method and performs equally good when predicting the state of the vehicle in a roundabout.

Englund~\citep{Englund2020a, Englund2020b} used real-world trajectories to predict the intention of cars or bicycles in an upcoming road exit. The AI methods used in~\citep{Englund2020a, Englund2020b} were based on Support Vector Machines~\citep{Vapnik1998}, Random Forest\citep{Breiman2001} and Multi-Layer Perceptrons\citep{Bishop1995a}.  In~\citep{Englund2020a} a backward elimination strategy for selecting the most important variables for predicting the behavior of the cars and bicycles in intersections was used. For bicycles the most important variables were speed, and for cars it was position. Heading was also among the 6 best variables for both vehicle types.


Garcia et. al.~\citep{garcia2017sensor} propose a mix of an Unscented Kalman Filter and Joint Probabilistic Data Association to fuse sensor reading from a vision-based system, a laser sensor and a global positioning system to obtain obstacle detection and object tracking. Li et. al.~\citep{li2013sensor} suggest combining LiDAR and vision-based sensors to obtain lane detection and extraction of an optimal drivable region. Sivaraman and Trivedi~\citep{sivaraman2013looking} review on-road vision-based vehicle detection, tracking, and behavior understanding. Beside Kalman filters, other algorithms such as Support Vector Machines~\citep{Vapnik1998}, Adaboost~\citep{freund1999short}, Hidden Markov Models~\citep{jazayeri2011vehicle} and Gaussian mixture modelling~\citep{wang2008automatic} are used for various fusion tasks. 
Recently, also deep learning in terms of Generative Adversarial Networks (GANs) has been used for sensor fusion of radar and camera sensor data.


In the context of multimodal  object detection, most of the recent works fuse RGB camera images with 3D LiDAR point clouds \citep{Frustum2017, Chen17, Liang2019CVPR}, whereas other works rather combine regular RGB data with thermal \citep{Takumi17}  or depth images \citep{Mees17}. In contrast to object detection, there are relatively few contributions related to multi-model semantic segmentation:  \citep{Valada19} fuses RGB, depth, and thermal images and  \citep{Kim18} combines RGB and LiDAR data streams for semantic segmentation.

One of the main challenges in multi-model perception is when to fuse various sensory readouts (e.g. RGB, LiDAR, etc.) which have vast variations in time scales, dimensions (i.e. 2D versus 3D data), and signal types (i.e. continuous versus discrete).
Deep neural networks, which are good at extracting and representing features hierarchically, provide various options to fuse sensor readings at different stages such as early, middle, and late. 
Early fusion \citep{Xu_2018_CVPR} directly merge raw data derived from different sensor modalities, e.g. first by concatenating raw scene features derived from different sensor modalities into a single vector and then training a deep neural network on this new feature representation.
Late fusion \citep{Wang19} combines learned unimodal sensor features at the highest network layer into a final prediction. Middle fusion \citep{Hazirbas2016} combines features learned at intermediate layers.

In contrast to other fusion strategies, early fusion requires less computation time and memory since the raw data readings are jointly processed. However, such methods are inflexible to changes in the network input type. For instance, when  a new sensing modality is introduced~\citep{tavares2016crowdsensing, alvear2018crowdsensing}, early fusion networks need to be retrained from scratch. In such cases, late fusion approaches are more flexible and ideal since only the domain-specific network needs to be retrained while the networks processing the other sensor data types remain the same. Although middle fusion networks are also relatively flexible, the network architecture design, i.e. finding the optimal combination of intermediate features, is non-trivial. 
Despite  having advanced fusion networks \citep{Xu_2018_CVPR,Wang19,Hazirbas2016} that achieve state-of-the-art performance on challenging object detection  and semantic segmentation datasets, lack of guidelines for designing optimal fusion networks still remains as a challenge since most networks are designed by empirical results \citep{Feng19}.

\subsection{Traffic System Control}\label{sec:control}

To plan the future transportation system, simulation tools are efficient tools. Besides giving realistic visualizations of the future transportation system, the simulation models can provide valuable information on how the system functions under different conditions. Simulation of Urban Mobility (SUMO) is a popular open source traffic micro simulation tool~\citep{krajzewicz2012recent,sumo2002}. It was used to simulate smart infrastructure that could improve traffic flow and energy efficiency~\citep{Englund2014}. One of the challenges with simulation is to validate the results. Building infrastructure is costly and the planning horizon is 50 years. To improve the simulation models, and to adapt to future vehicles that will have different level of automation, researchers at UC Berkeley have proposed a plug in for SUMO called Flow~\citep{kheterpal2018flow,wu2017flow}. Flow is developed to take into consideration both fully automated, semi-automated and manually driven vehicles into the simulation. The automated vehicle models take into consideration information from surrounding vehicles and infrastructure to be able to optimize the traffic behavior~\citep{wu2017flow}. 

A review on intersection management is given in~\citep{Chen2016a}. The paper discusses control strategies in signalized and non-signalized intersections. Four types of strategies have been investigated. Cooperative resource reservation concerns how vehicles reserve the tiles on their planned route for certain time slots to pass the intersection. Whereas resource allocation considers time slots and space tiles in for example intersections and roundabouts, trajectory planning concerns the scheduling of travel routes e.g. trajectory planning. Another strategy is to use virtual traffic lights to control traffic. The final approach is collision avoidance and is a complement to the above-mentioned ones. The resource planning or scheduling tools may have one plan, however the vehicle may have unknown constraints or deficiencies and therefore, collision avoidance for input control adjustments can be applied to make sure perpetual safety, e.g. collision avoidance in both short and long term. 

Graph Neural Networks (GNNs) have shown great potential to use existing traffic data to model future transportation systems and enable to perform counterfactual reasoning about factors that affect it. DeepMind in a collaboration with Google Map~\citep{deepmindGNN} has shown that the prediction of traffic and estimated arrival time improves by $50\%$ once the problem is formulated using GNNs. The graph represents the road structure and the artificial neural network learns the dynamics between roads building up the traffic system. The scalability of their approach enables modelling a complex structure of the traffic with a single model. Although GNNs have been around for several years, only today have they reached the maturity suitable for solving realistically complex problems, due to both algorithmic progress and new GPU-optimized implementations.

Forecasting any given parameter in the complex dynamics of traffic can be considered as a spatio-temporal problem. While spatial relations between road and road sections can be modelled with graph structure, the ways to model the temporal aspect can vary. Xie et al~\citep{Xie_2020} proposes a SeqGNN which combines sequence-to-sequence (Seq2Seq) models with GNNs. On the other hand, Song et al \citep{song-2018} models the temporal dependency between the graphs by recurrent approach and Guo et al \citep{guo-2019} adds an attention mechanism to control which weights need to be updated. Although RNN-based approaches seem to be a more popular choice for modelling the temporal aspects, Yu et al \citep{Yu_2018} propose a structure with several spatio-temporal convolutional blocks. The convolution structure has been defined on the time axis to model the temporal dependencies. Their goal is to exploit a simpler structure with fast training capabilities that can handle multi-scale traffic networks. 
Although a lot can be gained by modelling the roads and road connections as a graph, there is no guarantee that such a graph structure models the true underlying relationships between time-series. There might be a need for using graph embeddings \citep{chen2018}, proximity embedding \citep{Wang_2016}, and walk embedding \citep{Grover_2016} which can lower the dimensional complexity of the problem. A representation that is optimized to preserve both the proximity of nodes and the structure of the traversals is intuitively very appealing. In addition, the underlying graph might vary given different circumstances throughout time. Thus, Löwe et. al proposes to create amortized graphs \citep{lowe2020} that take advantage of the variation in the data.

\subsection{Driver monitoring}\label{sec:monitor}

With the introduction of more intelligent infrastructure and vehicles, one might conclude that the role that humans play will become less significant. 
However, some people might still want to drive themselves, which means that there will be a complex mixture of vehicles operating at different levels of autonomy (e.g.~in terms of the levels described by SAE International~\citep{SAE_LEVELS}).
Some potential dangers include ''risk compensation''--people engage in riskier behaviors because they think technology can deal with it--as well as lower driving ability due to less opportunities to drive.
Support can target accident-prone times, such as when a vehicle transitions to a more manual mode.
Moreover, when control is removed from a human driver, there is a responsibility to ensure that humans in a vehicle are comfortable and safe.
Thus, an important task includes detecting the state of people within a vehicle, to avoid negative states and seek to achieve positive states.
Negative states can involve sleepiness, distraction, drunkenness, health problems (e.g. epilepsy), and negative affect (anger, fear, and embarrassment), as well as individual predilections (some drivers can prefer to drive more wildly or be unsure how to interpret some driving situations due to lack of experience)~\citep{yang2020all}. Positive states can involve comfort and enjoyment~\citep{beggiato2020facial}.
To infer a human's state, some typical features detected include eye analysis (e.g. eye aspect ratio and blinking), as well as gaze, head pose, and posture.

In our previous work, we have explored the idea of using recurrent neural networks \citep{torstensson2019using} to estimate the future behavior of a car driver. A video data set was collected describing typical (future) driver behavior/activities, e.g. driving safe, glancing, leaning, removing hand from wheel, reaching, grabbing, retracting, and holding. A classification network was trained to recognize the current activity, and the result is fed into a recurrent network to predict the activity in the next frame. The accuracy of predicting the activity in the next frame is 80\%; nevertheless, the method is capable of predicting activity up to 20 frames ahead with an accuracy of 62\% (video was captured at 30fps). 
Furthermore, we have explored how social media could be used to support drivers, by leveraging insight into how they feel outside of the vehicle and interacting to reduce loneliness--a construct which has been tied to factors that increase the risk of accidents, such as depression and sleep deprivation~\citep{valle2021lonely}. %
Some open research challenges include how 'wisdom of the crowd' strategies can be incorporated to find potential dangers and anomalous driving, as well as how to infer the 'meaning' of detected emotions by detecting what they refer to: i.e. the 'emotional referent'. For example, if a passenger frowns, is this behavior a reaction to driving conditions, or to something on their cell phone?

Also, the need of continuous user authentication and monitoring is increasingly observable when larger fleet of professional vehicles are on the road and many drivers are brought under pressure to drive longer than what legislation allows. It can also aid for example to assess that an authorized person is driving a vehicle, or to detect driver drowsiness or distraction. 
Here, modalities captured with cameras (face \citep{GUO2019102805,9039580} or eye regions \citep{[Alonso18_perioc_expression],[Alonso16]}) can be complemented with sensors attached to the seat or the steering wheel that allows to capture bio-signals such as heartbeats \citep{Wartzek11} or skin impedance \citep{Macias13}, which correlates, for example, with sweating - stress level - but also with fitness levels \citep{Jaffrin08}.
There are also proof of concepts using Doppler radar for vital signs measurement \citep{Li13}, which has the evident advantage of not needing any type of contact.
Infrared thermal imaging is another possibility, derived from subcutaneous blood flow and perspiration patterns \citep{Ioannou14}, while allows to mitigate privacy concerns that may arise from the use of regular cameras operating in the visible range.
These solutions allow unobtrusive monitoring of human vital signs that goes beyond driver monitoring as well. 
While fatigue or distraction detection may seem the most straightforward task, other examples can include:
($i$) pre-crash road safety, since abnormal vital signs can reflect the presence of drugs, alcohol, stress, or even diseases such as pre-dementia \citep{Nicolini14};
($ii$) post-crash road safety, because detected signals can be used by Advance Automatic Crash Notification (AACN) systems in order to improve alarm handling; or
($iii$) person identity, which can be achieved in an unobtrusive way not only with traditional facial images, but also with bio-signals \citep{Maiorana18}.

Investments in driverless cars are already massive, both in the public and private sector. However, their benefits will be several orders of magnitudes higher if they are used collectively (taxis, buses, trucks feeding trains, etc.) reducing the number of vehicles serving transportation needs, producing higher time-gains and reducing environmental footprint. One bottle-neck will be then to secure the use of such vehicles when people who do not know each other travel together with more confidence and safety when there is no driver, and even without tickets, nor identity cards, since the identity and rights management can be done entirely by the system. %
In continuous biometrics, users are constantly monitored without needing active cooperation, in contrast to one-time authentication, e.g. at the beginning of a session. 
This may be done by using all pieces of biometric information available at a particular time, including soft-biometrics, behavior, or emotional state \citep{[Jain16]}. 
Accumulated evidence over time also can be used to improve accuracy. 
In this context, active modalities, such as fingerprints and iris, are often stronger than the weaker passive modalities (e.g. face), but the latter demand no cooperation. 
Additionally, certain intentions, expressions and physical states are noticeable in the footprint left only in continuous visual signals of face and body, e.g. drowsiness, stress, or irregular behavior.
In these scenarios, localization of body parts and handheld objects are also important for safety and comfort, in order to detect potentially dangerous people or events. For example, holding a book, a steering wheel or a weapon makes a big difference, as well as to who the hands belong or what the hands are engaged in.

In addition, for a vehicle with automated functions, knowing the driver to provide a pleasant experience is as important as knowing the actions and intentions of the surrounding road-users. In~\citep{varytimidis2018action} we developed AI-based algorithms that could detect the action and intentions of pedestrians. Such information is valuable while building reliable ADAS functions.

\section{Open research challenges and standardization}\label{sec:open_research}

The previous sections have outlined current research initiatives and state of the art within perception, traffic system control, and driver monitoring. We now explore the viable future research directions based on the previous work. Perception, from an ego vehicle is typically limited by the field of view of the available sensors which thus calls for reliable communication between road users and possible also infrastructure to improve local awareness and thereby allow for improved tactical and operational vehicle control. Standards such as Cooperative ITS~\citep{chencits2014} are put forward to ease collaboration in the traffic system. However, the current standard promotes simple \emph{here I am messages} and does not allow negotiation between road users to improve traffic safety and traffic flow. Future research should consider data formats and ontologies to enable interaction between not only vehicles and between vehicles and infrastructure but also between vehicles and VRUs. In addition, ontologies can be used to harmonize traffic behavior as described in~\citep{FASTZERO_2013}. Another challenge is the ownership of the intelligent infrastructure, the shared data and how to manage revenue. Consequently, to utilize the full potential of the technology, IoT and ICT, sharing of aggregated information such as trajectories or behavior should be enabled to further build local awareness and thus facilitate tactical decision-making in traffic.

Another area of future research is the security of the AI-based systems. Research should obstruct risks with sensor vulnerabilities. In e.g.~\citep{Anthony_Hack} the author highlights the risk of hijacking a vehicle with the help of manipulated billboards. Risks with hacking~\citep{ANDY_GREENBERG} is also prevalent and could remedy from time critical AI-based anomaly detection methods. Such methods would ensure safe and secure tactical and operational road vehicle automation.

As sensor technologies develop, and computing power increases the use of autonomous drones, both aerial and with wheels, will increase. With full anti-collision capabilities they will ease our lives with instant delivery, guiding,  carrying and surveillance. Such AI technology would need capabilities in all three domains, strategical, tactical and operational to be efficient. 

Another field that will evolve is how we interact with the technology. As mentioned in the introduction there have been attempts with light-based external HMI to ease the introduction of platoons in traffic, to let the surrounding traffic understand the intentions of the platooning vehicles. With higher level of automation, where the vehicles completely handle tactical and operational tasks, new ways of interacting with the vehicles are necessary. AI will play a major role to enable interaction with automated vehicles through gestures and speech. The vehicles can observe the current state of the driver or operator and thereby enable natural interaction and control in all three levels of automation, strategical, tactical and operational. 

With improved sensing and interaction, the introduction of unmanned vehicles will be expanded, both ground and aerial vehicles will become more and more common~\citep{ortiz2018uav}. 

Pervasive intelligence is another concept that can have great impact on both business and society. While systems become more and more digitized and start interacting more with other systems (trucks in a platoon or different administrations within a municipality or vehicles in a multi-modal transportation service system for example), there is always a risk of sub-optimization since the learning functions does not have access to the whole set of data. Consequently, future research should focus on how to optimize behavior on a larger scale, allowing the system to access data also outside of the own system boundaries.


One challenge of applying machine learning in vehicles is their rigorous safety requirements. The traditional standard ISO26262~\citep{ISO26262-1:2018} does not apply to a trained machine learning-based software since the behavior of a machine learning-based software is not explicitly expressed in source code and it does not follow a certain specification. The developers rather define an algorithm and an architecture that learns  the functionality. In the process enormous amounts of data complemented with domain-specific labels are used to teach the machine to capture the relationships between input and output data. For the development of road vehicle automation, this step usually concerns the collection and preprocessing of huge amount of data from e.g. camera, LiDAR, and radar sensors along with training and evaluation using even more data.

In January 2019, ISO/PAS 21448 - Safety of the Intended Functionality (SOTIF) was published containing guidance on the applicable design, verification and validation measures that are needed to achieve the SOTIF~\citep{ISOPAS21448:2019}. A PAS (Publicly Available Specification) is not an established standard, but rather a document that resembles the content of what is planned to be included in a future standard.

It is the intention that ISO 26262 and SOTIF should be complementary standards: ISO 26262 covers “absence of unreasonable risk due to hazards caused by malfunctioning behavior”~\citep{ISO26262-1:2018} by mandating rigorous development and is structured under the V-model way-of-working. The focus of SOTIF is to addresses “hazards resulting from functional inefficiencies of the intended functionality” ~\citep{ISOPAS21448:2019}, e.g. classification  failures in an automotive local awareness system, which is different from the type of malfunctions targeted by defect-oriented ISO 26262.

SOTIF is not structured based on the V-model but around ($i$) known safe states, ($ii$) known unsafe states, and ($iii$) unknown unsafe states. Note that SOTIF concerns the process of minimizing the two unsafe states, by focusing on detailing the requirements specifications for the developed functionality, where the aim is to  shift the hazards from ($iii$) $\rightarrow$ ($ii$) and from ($ii$) $\rightarrow$ ($i$) which in turn are derived from hazard identification and hazard mitigation, respectively.

Another challenge is harmonizing the introduction of more advanced vehicular functions and making them socially accepted. ADAS such as Forward vehicle collision mitigation systems  (FVCMS), Pedestrian detection and collision mitigation systems (PDCMS) and Bicyclist detection and collision mitigation systems (BDCMS) are examples of vehicle functions that make use of on-board sensors that build local awareness around vehicles. These systems are described and specified by  ISO,  International Organization for Standardization, i.e. FVCMS by ISO 22839:2013~\citep{ISO22839:2013}, PDCMS by ISO 19237:2017~\citep{ISO19237:2017} and BDCMS by ISO 22078:2020~\citep{ISO22078:2020}. These standards detail concepts of functionality, their minimum functionality and systems' requirements in conjunction with interfaces and how testing of the functions should be performed. Recently, standards on external Human-Machine Interfaces (eHMI) have been put forward. The functionality is described in the technical report~\emph{Road Vehicles — Ergonomic aspects of external visual communication from automated vehicles to other road users}, ISO/TR 23049~\citep{ISO/TR23049:2018}. The document describes how automated vehicles should communicate their intentions to surrounding road users. Enabling such functionality, the vehicles need to be able to interpret the behavior of their fellow road users.

\section{Summary and Conclusion}\label{sec:conclusion}
SCC refers to a cohesive concept to develop a sustainable future society. This paper highlights how AI can support the development of SCC and in particular within the future cooperative ITS. The two main challenges within ITS that are addressed by SCC are the traffic safety and the environmental challenge. In Europe, reducing the number of severely injured by half by 2030 from 2020 years level and the ultimate goal to have close to zero deaths in 2050 are two goals set by the European Commission~\citep{Commission2019}. 

    \textbf{Perception} using computer vision and sensing is one of the enablers of road vehicle automation. Sensor fusion is typically used to obtain a robust mapping of the surrounding. For the vehicle to understand, and interpret the surrounding, it uses semantic mapping that makes the vehicle aware. Another way to interpret the surrounding that is highlighted is to use classification and tracking to predict the behavior of surrounding road users.  
    
    \textbf{Traffic system control} is central in managing traffic flow in larger cities. To improve traffic system control, traffic simulators are often used. One can use simulators to do planning of cities or new road infrastructure as well as using them to predict future scenarios or the effect of a presumptive maintenance effort. Challenges include how to incorporate the effect of future automated vehicles, since their behavior is so far unknown.

    \textbf{Driver monitoring} is a research field that improves the user experience in vehicle automation. In ADAS the driver monitoring system helps to warn the driver if he/she becomes distracted or drowsy. The driver monitoring system also plays an important role in automated vehicles. The driver monitoring system should then know about the driver to be able to hand over control only when the driver is capable of driving. In-vehicle, sensors may also be used for authentication. For this purpose, camera sensors can be fused with other modalities, e.g. in the steering wheel or the seat, that can capture pulse or skin impedance. Doppler radar sensors are another example of system that with the help of AI can be used to estimate vital parameters of a driver, i.e. to improve handover or warning applications.

As reported in this paper, AI-based technology has achieved many great things and promises huge benefits in general concerning economic growth, social development, as well as quality of life and in particular reducing the environmental impact and improving traffic safety.

Most of the AI-based technologies that are mentioned in this paper are data intensive. Not only the volume and frequency of the data is important, but also there is a need to merge a variety of data from different types and sources. Data from service providers, municipalities, traffic authorities, and user data are required to create a holistic view of the situation on a scale of a city. Collecting and analyzing such data comes with privacy challenges such as a trade-off between preserving the rights of the road users and providing personalized services. The possibilities of decentralized analysis of data or aggregation of intermediate results could also be considered.

The common challenge in real world applications of AI is how to trust the system--since the behavior of an AI is not built by explicitly expressed source code and does not follow a certain specification but is rather built by algorithms that learn the intended functionality based on historical data. The generalizability of the algorithms in a real-world setting can be measured when they are deployed in practice and they are forced to encounter corner cases that have not been encounter before.

However, one of the main challenges is how can we set up the requirements of a system that is based on historical data. In road vehicle automation, where safety is imperative, verification and validation are crucial to benefit from the generalizability of the AI technology. Even if the training data are annotated and contain labels of the objects in the scenes, what are the guarantees that no new objects, different from the ones in the training set, will appear in future traffic situations? In addition, what is the relationship between requirements and the level of detail of the data annotation?

In addition, the low-level explainability of AI models, as well as data biases and data privacy pose considerable risks for users, developers, humanity, and societies. Consequently, although AI can predict, model and sense, AI technology is not yet capable of explaining how it comes to certain conclusions and therefore, it is difficult for humans to fully trust it.

All in all, enabling technologies for smart transport has come a long way in recent years. Now, it is time to start to connect these technological building blocks to unlock socio-economic benefits of smart cities. This step is going to be challenging and it needs a collaboration of several actors from end-users, SMEs, Original Equipment Manufacturer (OEMs), city responsible, legislators to implement existing technologies in practice.

\bibliographystyle{chicago}
\spacingset{1}
\bibliography{references}
	
\end{document}